\documentclass[journal]{IEEEtran}
\usepackage{amsmath,amsfonts}
\usepackage{algorithmic}
\usepackage{algorithm}
\usepackage{array}
\usepackage[caption=false,font=normalsize,labelfont=sf,textfont=sf]{subfig}
\usepackage{textcomp}
\usepackage{stfloats}
\usepackage{url}
\usepackage{verbatim}
\usepackage{mathrsfs}
\usepackage{cite}
\usepackage{graphicx}
\usepackage{bbding}
\newcommand{\tabincell}[2]{\begin{tabular}{@{}#1@{}}#2\end{tabular}}  
\usepackage{xcolor, colortbl}
\usepackage{times}
\usepackage{epsfig}
\usepackage{graphicx}
\usepackage{amsmath}
\usepackage{amssymb}
\usepackage{bm}
\usepackage{booktabs}
\usepackage{setspace}
\usepackage{float}
\usepackage{color}
\usepackage{multirow}
\definecolor{cyan}{cmyk}{.3,0,0,0}
\usepackage{threeparttable}
\usepackage[colorlinks=true,citecolor=green]{hyperref}
\newcommand{\figref}[1]{Fig.~\ref{#1}}%
\newcommand{\tabref}[1]{Tab.~\ref{#1}}%
\newcommand{\secref}[1]{Sec.~\ref{#1}}
\renewcommand{\eqref}[1]{Eq.~(\ref{#1})}

\newcommand{\etal}{\textit{et al. }}
\newcommand{\ie}{\textit{i.e.}}

\newcommand{\eg}{\textit{e.g.}}

\newcommand{\vs}{\textit{vs.}}

 \usepackage[switch]{lineno}

\begin{document}
\title{GaTector+: A Unified Head-free Framework for Gaze Object and Gaze Following Prediction}

\author{Yang Jin,
        Guangyu Guo,
        Binglu Wang
 \thanks{Yang Jin, Guangyu Guo and Binglu Wang are with the School of Astronautics, Northwestern Polytechnical University, Xi'an, China. E-mail: jin91999@gami.com; gyguo95@gmail.com; wbl921129@gmail.com (Corresponding author: Binglu Wang).}
}


\IEEEtitleabstractindextext{%
\begin{abstract}
Gaze object detection and gaze following are fundamental tasks for interpreting human gaze behavior or intent. However, most previous methods usually solve these two tasks separately, and their prediction of gaze objects and gaze following typically depend on head-related prior knowledge during both the training phase and real-world deployment. This dependency necessitates an auxiliary network to extract head location, thus precluding joint optimization across the entire system and constraining the practical applicability. To this end, we propose GaTector+, a unified framework for gaze object detection and gaze following, which eliminates the dependence on the head-related priors during inference. Specifically, GaTector+ uses an expanded specific-general-specific feature extractor that leverages a shared backbone, which extracts general features for gaze following and object detection using the shared backbone while using specific blocks before and after the shared backbone to better consider the specificity of each sub-task. To obtain head-related knowledge without prior information, we first embed a head detection branch to predict the head of each person. Then, before regressing the gaze point, a head-based attention mechanism is proposed to fuse the sense feature and gaze feature with the help of head location. Since the suboptimization of the gaze point heatmap leads to the performance bottleneck, we propose an attention supervision mechanism to accelerate the learning of the gaze heatmap. Finally, we propose a novel evaluation metric, mean Similarity over Candidates (mSoC), for gaze object detection, which is more sensitive to variations between bounding boxes.
The experimental results on multiple benchmark datasets demonstrate the effectiveness of our model in both gaze object detection and gaze following tasks.
\end{abstract}


}
\maketitle

\IEEEdisplaynontitleabstractindextext
\IEEEpeerreviewmaketitle

\section{Introduction}
\label{sec:introduction}

The objects stared at by humans have vital value for analyzing human gaze behavior, which can potentially reveal a person's state of mind and behavioral intentions. Therefore, as an important aspect of understanding human gaze behavior, \textit{Gaze Following}~\cite{gupta2022modular, tonini2022multimodal, miao2023patch, jin2021multi, chen2021gaze, hu2022gaze, recasens2015they, recasens2017following,chong2018connecting,chong2020detecting,lian2018believe, Tu_2022_CVPR,bao2022escnet,fang2021dual, wang2023dualgaze} is intended to infer the focus of human attention, which also plays an important role in interpreting human behavior due to its perception and understanding of the surrounding environment. Taking into account visual focus while predicting the objects gazed at by humans can further enhance the human-object and human-scene semantic connection and provide a richer understanding of human gaze behavior, thereby expanding the application scope of gaze behavior analysis. Similarly, \textit{Gaze Object Detection}~\cite{tomas2021goo, wang2022gatector}, which aims to detect the location and category of the object captured by human gaze behavior, can be widely used in various scenarios in the real world. For example, a person looking at their watch in front of a bus station may indicate that they have something urgent to do. Likewise, a customer staring at a product in a shopping mall may be interested in purchasing it~\cite{bermejo2020eyeshopper}. 




\begin{figure*}[!t]
\centering
\includegraphics[width=1\linewidth]{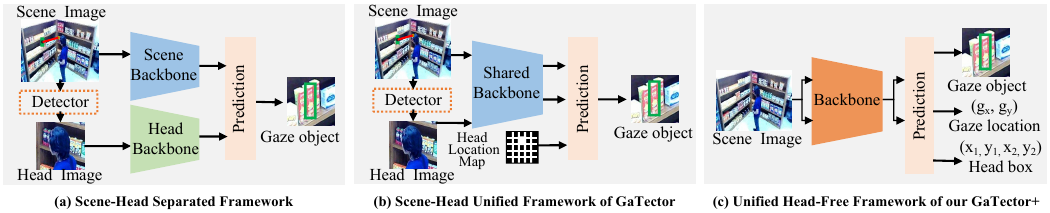}

\caption{Comparison of different strategies for gaze object prediction. (a) Two separate branches are used to extract head and scene features. (b) GaTector~\cite{wang2022gatector} uses a unified framework to solve this problem but still requires a head image as input in the inference process. (c) GaTector+ does not need any auxiliary detector and head prior information during the inference.}
\vspace{-0.45cm}
\label{fig:intro}
\end{figure*}

Many works have explored gaze following~\cite{gupta2022modular, tonini2022multimodal, miao2023patch, jin2021multi, chen2021gaze, hu2022gaze, recasens2015they, recasens2017following,chong2018connecting,chong2020detecting,lian2018believe, Tu_2022_CVPR,bao2022escnet,fang2021dual, wang2023dualgaze, tu2023gaze, tonini2023object}. However, as a recently proposed task, gaze object detection has rarely been explored. Tomas \etal \cite{tomas2021goo} proposed this task and released the first two benchmark datasets, \ie, GOO-Synth and GOO-Real for this task, but did not implement a specific solution. As shown in~\figref{fig:intro} (a), a simple way to realize gaze object detection is to follow previous gaze following methods and use two separated networks to extract features from the scene image and head image, but this mechanism prevents each network from joint optimization and obstructs information fusion. More recently, we proposed a unified framework for gaze object detection called GaTector~\cite{wang2022gatector} (as shown in ~\figref{fig:intro} (c)), which uses a specific-general-specific (SGS) feature extractor with a shared backbone to extract general features from the scene and head images. This framework achieves efficient gaze object detection while maintaining a lightweight structure.

However, previous methods, whether for gaze object detection or gaze following tasks, both require additional head-prior information as input, leading to the need for additional head auxiliary networks during inference, which affects the practicality and flexibility of the model in a wider range of applications. Even though some gaze following works~\cite{tu2023gaze, Tu_2022_CVPR} perform global modeling by using Transformer~\cite{NIPS2017_3f5ee243} to solve this problem, they lack lack understanding of specific head information, resulting in relatively poor results. Furthermore, the existing methods solve gaze object detection or gaze following separately, which is either a lack of understanding of scene interaction or a lack of understanding of object details.



To this end, this paper builds a unified head-free framework for gaze object detection and gaze following, which detects gaze objects while taking into account the prediction of visual attention focus and does not require any head-related prior information during the inference. Specifically, to establish a connection between scene information and gaze information in a unified framework without head-prior information, we extend the SGS mechanism and extract head-specific features from the scene image through a head-specific block, which is named SGS+. Then, we propose a head-free gaze object and gaze following predictor, which decodes visual focus while detecting gaze objects. For this module, to predict all individuals' corresponding gaze objects and visual focus in an image, we first embed a head branch based on the head-specific features. Then a gaze attention module based on the gaze-specific feature and predicted head location map is proposed to generate a gaze attention map that can introduce gaze contexts to the scene-specific features. Since the coarse head location map during training causes inaccurate attention maps to be generated, an attention supervision mechanism is introduced to correct the attention maps and accelerate the learning of the gaze regressor, which further enhances visual attention in the gaze following task while enabling more accurate gaze object localization. At the end of the model, a box energy aggregation loss is introduced to guide the gaze heatmap to focus on the stared box and improve the gaze following accuracy.

Traditionally, evaluation metrics for object detection tasks have mainly been based on Intersection over Union (IoU). However, IoU has limitations when it comes to evaluating gaze object detection (GOD) since it only considers spatial overlap between predicted and ground-truth objects, and cannot make accurate measurements when two boxes do not have overlapping areas. Although Gatector proposed a weighted Union over Closure (wUoC) metric~\cite{wang2022gatector} that reflects the distance between the real and detected objects, it will fail when the area of the predicted box and the ground truth box are equal and adjoin each other (shown in Fig.~\ref{fig:msoc} (b)). Therefore, to accurately measure the performance of the GOD task, we propose a new evaluation metric called mean Similarity over Candidates (mSoC), which solves the above limitations by refining the weights and is more robust in evaluating GOD tasks.

In summary, the contributions of our paper are mainly reflected in the following four aspects: 
\begin{enumerate}
\item We introduce the GaTector+, a unified head-free framework for gaze object detection and gaze following, which eliminates dependence on head-prior information in the inference, and enriches the semantic understanding of human gaze behavior in various scenarios while enhancing the practicability and flexibility in practical applications.
\item To break through the performance bottleneck caused by gaze regression, a head-based attention module is proposed to fuse the sense feature and gaze feature, and an attention supervision mechanism is introduced to accelerate the learning of the gaze heatmap.
\item We propose a new metric mSoC, which can address the limitations of the previous wUoC metric and is more robust for evaluating the gaze object detection task. 
\item Extensive experiments on multiple datasets demonstrate the effectiveness of our model on gaze object detection and gaze following tasks when head prior is not provided, thereby providing a richer semantic explanation for human gaze behavior.
\end{enumerate}

\section{Related Works}
\label{sec:relat}

\subsection{Gaze Object Detection}
\label{sec:relat_gop}
As a recently proposed task, gaze object detection~\cite{tomas2021goo,wang2022gatector} is rarely explored. Unlike simple motion prediction and localization~\cite{10948357,9171561,9633199,9794923,9880381}, gaze object detection intend to interpret the gaze interaction relationship between people and objects in the image. Tomas~\etal~\cite{tomas2021goo} first proposed the gaze object detection task and released a benchmark dataset, GOO. Similar to gaze target prediction, gaze object detection intends to predict the gaze object that the target person is looking at. The difference is that while gaze target prediction only predicts the gaze point where the target person is looking, gaze object detection predicts both the gaze point and the gaze object bounding box and category. Gaze object detection is an essential clue for understanding human intention located in a picture. Wang~\etal proposed a Specific-General-Specific framework called GaTector~\cite{wang2022gatector} to solve the gaze object detection. GaTector reduced the number of parameters by using a shared backbone that simultaneously extracts the scene and head features. Although GaTector provides a unified solution for gaze object detection, the additional head-prior information hinders its application in real scenarios. In contrast, the proposed GaTector+ and SGS+ feature extraction mechanism only requires the scene image containing the target person as input, making the practical application of our model possible.

\subsection{Gaze Following}
\label{sec:relat_ge}
Earlier works on gaze are mainly divided into two categories: model-based~\cite{guo2021hybridgazenet, huang2023gaze} and appearance-based approaches~\cite{zhang2015appearance, Zhang_2022_CVPR,leifman2017learning, li2022appearance}. For the model-based methods, detailed eye features are first detected and then a geometric model is used to obtain the gaze direction~\cite{jin2023redirtrans, cai2023source}. Appearance-based methods based on neural networks encode the full-face image to predict the gaze direction~\cite{zhang2015appearance, cheng2022gaze, Cheng_2023_ICCV,cheng2024appearance,cheng2024you,yin2024clip,wu2023attention,fu2020purifying, kuric2025democratizing, nieva2025towards}. However, both approaches only predict the gaze direction and ignore the gaze target. To address this limitation, some studies focus on predicting the gaze following, which is the point that a person is looking at~\cite{gupta2022modular, tonini2022multimodal, miao2023patch, jin2021multi, chen2021gaze, hu2022gaze, recasens2015they, recasens2017following,chong2018connecting,chong2020detecting,lian2018believe,Tu_2022_CVPR,bao2022escnet,fang2021dual,Hu_2023_CVPR,horanyi2023they, tonini2023object,tu2023gaze, wang2023dualgaze, yang2024gaze,yang2024gazeecc, tafasca2024sharingan, tafascatoward}. Recasens~\etal\cite{recasens2015they} have built a baseline model on the GazeFollowing dataset for gaze following. Chong \etal \cite{chong2018connecting,chong2020detecting} released the VideoAttentionTarget benchmark dataset for visual attention detection in videos. Gupta \etal \cite{gupta2022modular} proposed a modular multimodal architecture for gaze target prediction that takes privacy into account. Fang \etal \cite{fang2021dual} achieved comparable performance by incorporating 3D gaze direction prediction into gaze target prediction. Removing the head image has been explored to improve the practicality of gaze target prediction in recent studies. Tu \etal \cite{Tu_2022_CVPR, tu2023gaze} proposed an end-to-end model that simultaneously predicts the head location of all individuals and their gaze locations. Recently, Wang~\etal\cite{wang2023dualgaze} proposed a dual regression-enhanced gaze position prediction method to achieve direct regression of 2D coordinates. However, the gaze following task only serves the prediction of visual focus and does not consider the gazed object, which limits their application scenarios. Therefore, we proposed to extend the scope of gaze prediction to include predicting the gazed object and visual focus directly from the scene image containing the target person, making our model more applicable to real-world scenarios.

\subsection{Object Detection}
\label{sec:relat_od}
Object detection~\cite{Law_2018_ECCV,Duan_2019_ICCV,Zhu_2019_CVPR,Tian_2019_ICCV,duan2020corner,zhu2020soft, lin20223d,Girshick_2014_CVPR,Girshick_2015_ICCV,NIPS2015_14bfa6bb,Cai_2018_CVPR,Lu_2019_CVPR,redmon2016you, li2025multi, chen2025hybrid, liu2025small} is an essential computer vision problem that involves identifying all objects of interest in an image and determining their categories and positions. Anchor-free~\cite{Law_2018_ECCV,Duan_2019_ICCV,Zhu_2019_CVPR,Tian_2019_ICCV,duan2020corner,zhu2020soft, lin20223d} and anchor-based~\cite{Girshick_2014_CVPR,Girshick_2015_ICCV,NIPS2015_14bfa6bb,Cai_2018_CVPR,Lu_2019_CVPR,redmon2016you} methods are commonly used in object detection. Anchor-free methods detect the upper left and lower right corners of the object and directly detect the central region and boundary information of the object. On the other hand, anchor-based methods can be further divided into two-stage and one-stage methods. Two-stage methods~\cite{Girshick_2015_ICCV,NIPS2015_14bfa6bb, guo2017video} first extract features of the object region and then classify and identify the region. In contrast, one-stage methods directly predict the bounding box and category confidence based on the extracted feature map. Zhao \etal \cite{zhao2019object} made a general review on object detection based on deep learning. Although transformer-based object detection~\cite{zhu2020deformable,meng2021conditional,liu2022dab,wang2022anchor,zhang2022dino,liu2022dab,li2022dn} has achieved promising results, its reasoning structure is relatively complex and the global modeling mechanism is not friendly to small object detection. For the gaze object detection task, one-stage CNN-based methods like YOLO~\cite{redmon2016you, chengyolo2024}, its local receptive field mechanism is relatively friendly to small objects and has real-time performance. Therefore, in GaTector, YOLOv4~\cite{bochkovskiy2020yolov4} is used as an object detector to detect objects that may be gazed at. Our proposed GaTector+ also uses the same detector for a fair comparison.

\begin{figure*}[!t]
\centering
\includegraphics[width=1\linewidth]{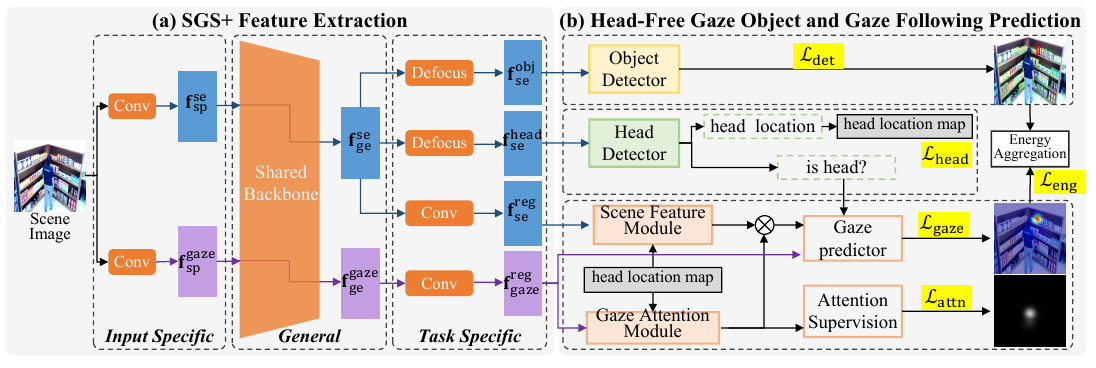}

\caption{Overview of the proposed GaTector+. (a) The specific-general-specific mechanism (SGS+) consists of a sense-specific branch and a gaze-specific branch while sharing the backbone to fuse information. (b) The head-free gaze object and gaze location prediction consist of an object detector, head detector, and gaze regressor. The gaze regressor has a head location-based attention mechanism to fuse the sense and gaze information, we propose an attention supervision mechanism to reduce the learning difficulty of the gaze regressor process. An energy aggregation loss is used to jointly optimize the object detector and gaze regressor.}
\vspace{-0.45cm}
\label{fig:framework}
\end{figure*}

\section{Method}
\label{sec:method}

Given an image, our goal is to take into account visual focus prediction while detecting the bounding boxes and categories of objects gazed at by humans. In this section, we first present the overall framework of our GaTector+, and then we introduce the extended specific-general-specific feature extraction mechanism (SGS+), object detector, head detector, gaze regressor, and box aggregation energy loss in detail. Finally, we give a detailed introduction to the proposed mSoC metric.

\subsection{Overview}
\label{sec:method_overview}
In this paper, we aim to propose a unified head-free framework for gaze object detection and gaze following to enhance understanding of human gaze behavior and achieve practical application. As illustrated in~\figref{fig:framework}, GaTector+ only requires a scene image as input to detect objects and regress gaze heatmap during the inference. GaTector+ consists of an SGS+ feature extractor and a head-free gaze object and gaze location prediction network. The SGS+ feature extractor only takes the scene image as input to generate task-specific features. The head-free gaze object and gaze location prediction network takes task-specific features as input and predicts all objects and the gaze heatmap.

To effectively extract features for head-free gaze object detection and gaze following, we extend the SGS mechanism to SGS+. SGS+ maintains the specific-general-specific structure of SGS in GaTector. Two input-specific blocks are used to extract scene-specific features and gaze-specific features, respectively. But in SGS+, two input-specific blocks both take scene images as input. Then, a shared backbone is used to extract general features. Finally, several task-specific blocks are used to extract task-specific features. Different from GaTector~\cite{wang2022gatector}, a new task-specific block is proposed to produce features for the head detector. The details of the SGS+ mechanism are introduced in~\secref{sec:method_sgs+}.

The head-free gaze object and gaze location prediction network consist of an object detector, a head detector, and a gaze regressor.
The \textbf{object detector}, which is introduced in~\secref{sec:method_od}, aims to precisely predict locations and categories for all candidate gaze objects in the retail scenario. The loss of the object detector and \textbf{head detector} is denoted by $\mathcal{L}_{det}$ and $\mathcal{L}_{head}$, respectively. The \textbf{head detector} is proposed to detect the head location and provide a head location map $\mathbf{H}$ that is useful in the gaze regressor. The head detector is introduced in ~\secref{sec:method_head_detector}. The \textbf{gaze regressor}, which is introduced in ~\secref{sec:method_ge}, aims to predict gaze heatmaps for all gaze objects and decode the visual focus. Since GaTector+ does not have a precise head location during training, we propose a head-based gaze attention module to generate an attention map with the help of the head detector and an attention supervision mechanism, which can generate more precise attention weights for the gaze regressor. 
The attention map and the gaze heatmap are optimized by attention loss $\mathcal{L}_{attn}$ and the gaze regression loss $\mathcal{L}_{gaze}$, respectively. Besides the aforementioned optimization losses for each task, an energy aggregation loss $\mathcal{L}_{eng}$ is used to jointly train the object detector and the gaze regressor.

In the training process, the overall loss function is defined as:
\begin{equation}
    \mathcal{L} = \mathcal{L}_{det} + \mathcal{L}_{head} + \alpha\mathcal{L}_{attn} + \beta\mathcal{L}_{gaze} + \gamma\mathcal{L}_{eng}.
    \label{eqtotalloss}
\end{equation}
where $\alpha$, $\beta$, and $\gamma$ are the weights of attention loss $\mathcal{L}_{attn}$, the gaze regression loss $\mathcal{L}_{gaze}$, and the box energy aggregation loss $\mathcal{L}_{eng}$, respectively.

\subsection{SGS+ Feature Extraction}
\label{sec:method_sgs+}
GaTector~\cite{wang2022gatector} proposes a specific-general-specific (SGS) feature extraction mechanism, which first utilizes two input-specific blocks for the scene image and head image, respectively, and then a shared backbone is used to extract general features for object detection and gaze estimation. Finally, SGS uses output-specific blocks to extract task-specific features. However, SGS does not apply to the head-free gaze object detection task since it requires a specific head image as input. To address this limitation and maintain the advantages of the SGS feature extraction mechanism, we introduce the SGS+ feature extraction mechanism for head-free gaze object detection. 

As shown in ~\figref{fig:framework} (a), our SGS+ feature extraction mechanism consists of the input-specific feature extractor, the general feature extractor, and the task-specific feature extractor. Firstly, the input-specific feature extractor utilizes two independent convolutional layers to extract the scene-specific feature and the gaze-specific feature from the scene image:
\begin{equation}
  \mathbf{f}_{\rm sp}^{\rm se} = \psi^{\rm se} (\mathbf{I}_{\rm se}), \ \ \mathbf{f}_{\rm sp}^{\rm gaze} = \psi^{\rm gaze} (\mathbf{I}_{\rm se}),
\end{equation}
where $\psi^{\rm se}(\cdot)$ and $\psi^{\rm gaze}(\cdot)$ denote independent convolutional layers. $\mathbf{f}_{\rm sp}^{\rm se}$ and $\mathbf{f}_{\rm sp}^{\rm gaze}$ denotes the scene-specific feature and gaze-specific feature.

Then $\mathbf{f}_{\rm sp}^{\rm se}$ and $\mathbf{f}_{\rm sp}^{\rm gaze}$ are fed into the general feature extractor to extract the general features by a shared backbone:
\begin{equation}
  \mathbf{f}_{\rm ge}^{\rm se} = \psi^{\rm ge} (\mathbf{f}_{\rm sp}^{\rm se}), \ \ \mathbf{f}_{\rm ge}^{\rm gaze} = \psi^{\rm ge} (\mathbf{f}_{\rm sp}^{\rm gaze})
\end{equation}
where $\mathbf{f}_{\rm ge}^{\rm se}$ denotes the scene-general feature, $\mathbf{f}_{\rm ge}^{\rm gaze}$ denotes the gaze-general feature, and $\psi^{\rm ge}$ denotes the shared backbone. By sharing the backbone, the network parameters would be greatly reduced, and the scene and gaze features and be better fused to build a basic relationship between the detection task and gaze regression task.

Afterward, in the task-specific feature extractor, we take the scene-general feature $\mathbf{f}_{\rm ge}^{\rm se}$ as input and use the \textit{Defocus} layer (see in ~\secref{sec:method_od}) to generate object-specific feature $\mathbf{f}_{\rm se}^{\rm obj}$ and head-specific feature $\mathbf{f}_{\rm se}^{\rm head}$ for detecting object and head location:
\begin{equation}
  \mathbf{f}_{\rm se}^{\rm obj} = \phi_{\rm df}^{\rm obj} (\mathbf{f}_{\rm ge}^{\rm se}), \ \
  \mathbf{f}_{\rm se}^{\rm head} = \phi_{\rm df}^{\rm gaze} (\mathbf{f}_{\rm ge}^{\rm se}),
\end{equation}
where $\phi_{\rm df}^{\rm obj}$ and $\phi_{\rm df}^{\rm gaze}$ denotes two \textit{Defocus} layers. Then, both scene-general feature $\mathbf{f}_{\rm ge}^{\rm se}$ and gaze-general feature $\mathbf{f}_{\rm ge}^{\rm gaze}$ are sent into two independent convolutional layers to generate gaze regression specific features:
\begin{equation}
  \mathbf{f}_{\rm se}^{\rm reg} = \phi^{\rm se} (\mathbf{f}_{\rm ge}^{\rm se}), \ \ \mathbf{f}_{\rm gaze}^{\rm reg} = \phi^{\rm gaze} (\mathbf{f}_{\rm ge}^{\rm gaze}),
\end{equation}
where $\phi^{\rm se}(\cdot)$ and $\phi^{\rm gaze}(\cdot)$ indicate two independent convolution layers. $\mathbf{f}_{\rm se}^{\rm reg}$ and $\mathbf{f}_{\rm gaze}^{\rm reg}$ indicate the scene-specific-regression feature and gaze-specific-regression feature, respectively. Then, $\mathbf{f}_{\rm se}^{\rm reg}$ and $\mathbf{f}_{\rm gaze}^{\rm reg}$ will be fed into the gaze regressor to obtain an accurate gaze heatmap. (see in~\secref{sec:method_ge}).

The SGS+ feature extraction mechanism provides an effective solution to feature extraction for head-free gaze object detection gaze following tasks, which allows us to use scene images to extract detailed gaze-specific features in the absence of head images while retaining the original advantages of the SGS. Compared with SGS,  we make two modifications: \textbf{(1)} To get the head location during training and inference, we add a head-specific defocus layer to extract the head-specific features. \textbf{(2)} Since the head-free gaze object detection and gaze following do not have the head image as input, we use the scene image to extract gaze-specific features instead.

\begin{figure}[!t]
\includegraphics[width=0.75\linewidth]{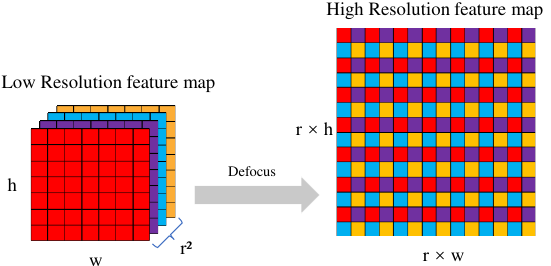}
\caption{Illustration of the \textit{Defocus} operation. One channel in the high-resolution feature map is transformed from $r^{2}$ channels in the low-resolution feature map.}
\vspace{-0.45cm}
\label{fig:ps}
\end{figure}

\subsection{Defocus Layer and Object Detector}
\label{sec:method_od}
In this paper, we follow GaTector~\cite{wang2022gatector} to use YOLOv4\cite{bochkovskiy2020yolov4} as our object detector to detect all potential objects of interest in the retail environment for a fair comparison. 

\noindent \textbf{\textit{Defocus} layer.} Since the retail scene comprises many small objects, high-resolution features are needed. A simple way is to use interpolation operations to enlarge the feature map $\mathbf{f}_{\rm ge}^{\rm se} \in \mathbb{R}^{2048 \times 7 \times 7}$, resulting in a high-resolution feature map $\mathbf{f}_{\rm ge}^{\rm se}{'} \in \mathbb{R}^{2048 \times 14 \times 14}$. While this strategy significantly improved the detection performance, it also increased the computational costs of the model. In order to solve the extra computational cost brought by interpolation, we use the \textit{Defocus} layer, which enables us to enlarge the feature map without compromising information and incurring additional computational costs. Specifically, in GaTector, our \textit{Defocus} layer uses the pixel shuffle~\cite{shi2016real} to enlarge the feature map from the last three layers of ResNet-50. As depicted in ~\figref{fig:ps}, the Defocus layer rearranges the feature map elements and reduces the channel dimension by a factor of $1/r^2$ for a given enlargement ratio $r$. Subsequently, the height and width of the feature map are magnified by a factor of $r$. This rearrangement process is computationally efficient and retains all the information in the object-specific feature $\mathbf{f}_{\rm ge}^{\rm se}$.

\noindent \textbf{Object detector.}
Based on the scene-general feature, we utilize the \textit{Defocus} layer to generate object-specific feature maps $\mathbf{f}_{\rm se}^{\rm obj}$ and then feed $\mathbf{f}_{\rm se}^{\rm obj}$ into the object detection module to obtain the prediction results. Furthermore, due to the superior performance of the \textit{Defocus} layer, we have replaced the interpolation operation in the detection head with the \textit{Defocus} layer, which requires fewer computational resources but delivers precise detection results. Since the size of objects in the retail scene remains relatively constant, we directly employ a high-resolution feature map to detect the objects and eliminate other detection heads, further reducing the computational cost of the model. The loss of object detector $\mathcal{L}_{\rm det}$ consists of a classification loss, an overlapping loss, and a box regression loss. Following the setting of GaTector, we employ the multi-class cross-entropy loss as the classification loss to supervise the accuracy of categories predicted by the object detector and employ the binary cross-entropy loss as the overlapping loss. For the box regression loss, we adopt the CIoU loss~\cite{zheng2020distance} to increase the predicted bounding box size and aspect ratio error.

\subsection{Head detector}
\label{sec:method_head_detector}
Previous methods~\cite{recasens2015they, lian2018believe, chong2020detecting, wang2022gatector} has demonstrated that the head image and head location map can provide valuable guidance for the gaze following and localization of gaze objects. To incorporate head location information in the head-free gaze object detection network, we propose the head detector to detect the head location and generate the head location map. In the head detector, we take the head-specific feature $\mathbf{f}_{\rm se}^{\rm head}$ as the input. For each anchor, we utilize a classification head to provide the head confidence score and another regressor head to predict the head bounding box. Finally, we generate the head location map based on the head location. The head detector loss consists of the head box regression loss and the confidence loss. 
\begin{equation}
    \mathcal{L}_{\rm head} = \mathcal{L}_{\rm head}^{\rm reg} + \mathcal{L}_{\rm head}^{\rm conf}.
\end{equation}

\noindent\textbf{Is head?} During the training, we used the ground truth box to calculate the IoU for each anchor. For each ground truth box, we search for an anchor box with the maximum IoU as the prediction box. If the IoU between the prediction box and the ground truth box is greater than 0.5, we use the prediction box to generate the corresponding head location map for the gaze regressor. When testing, we first decode the output from the head detector and then use non-maximum suppression to filter out redundant head bounding boxes to get the final head detection result. Then, we use the generated head location map to obtain the heatmap corresponding to each head. 
We use binary cross-entropy to calculate the head bounding box classification loss during the training:
\begin{equation}
    \mathcal{L}_{\rm cls}^{\rm head} =  - [h_{\rm log}\hat{h} + (1 - h){\rm log}(1 - \hat{h})],
\end{equation}
where $\hat{h}$ indicates the head confidence after the sigmoid activation.

\noindent \textbf{Head location regressor \& head location map.}
Similar to GaTector, we use the same box regressor as YOLOv4\cite{bochkovskiy2020yolov4} in the head detector and adopt the CIoU loss~\cite{zheng2020distance} to train the Head location regressor. During training, if the IoU between the predicted head box and ground truth is larger than 0.5, we use the predicted head bounding box coordinates to generate a head location map $\mathbf{H}\in \mathbb{R}^{H_{\rm T} \times W_{\rm T}}$. The values that we used are $H_{\rm T}, W_{\rm T}=224$. Specifically, we first generate a matrix and initialize each value of the matrix to 0, then we calculate the values in the matrix by:
\begin{equation}
    \mathbf{H}_{i,j} = 
    \begin{cases}
        1, & \text{if}\  \text{$x_1 \leq i  \leq x_2$ and $y_1 
 \leq j  \leq y_2$} \\
        0, & \text{otherwise}
    \end{cases}
\end{equation}
where $(x_1,y_1),(x_2,y_2)$ represent the coordinates of the upper left corner and the lower right corner of the head bounding box, respectively. $\mathbf{H}_{i,j}$ represents each value calculated of the matrix $\mathbf{H}$. The head location map $\mathbf{H}$ is used to provide the head location information for the gaze regressor.

\begin{figure}[!t]
\centering
\includegraphics[width=1\linewidth]{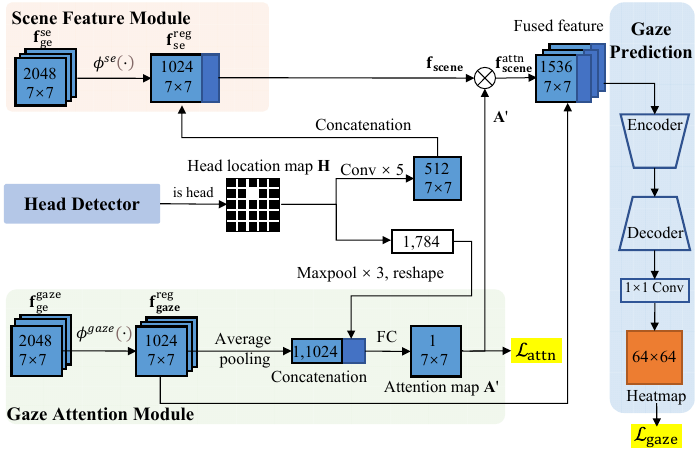}
\caption{Illustration of the gaze regressor.}
\vspace{-0.45cm}
\label{fig:fusion}
\end{figure}

\subsection{Gaze regressor}
\label{sec:method_ge}
The gaze regressor is intended to regress an accurate gaze heatmap for the gaze object while providing a visual focus for gaze following tasks. As illustrated in Fig.~\ref{fig:fusion}, gaze regressor uses two regression-specific features $\mathbf{f}_{\rm se}^{\rm reg}$ and $\mathbf{f}_{\rm gaze}^{\rm reg}$ as the input, and uses the head location map to improve the accuracy of gaze prediction. The gaze regressor consists of a scene feature module, a gaze attention module and a gaze prediction module.

\noindent\textbf{Scene feature module.}
The scene feature module processes the head location map $\mathbf{H}$ by five convolutional layers. Then, we concatenate the output with the scene-regression-specific features $\mathbf{f}_{\rm se}^{\rm reg}$ to make the feature map aware of the head location and generate the coarse-grained feature that has a coarse-grained connection between target person and object. The output of the scene feature module is denoted as $\mathbf{f}_{\rm scene}$.

\noindent\textbf{Gaze attention module.} In this paper, we use a head-based gaze attention module to generate an attention map that focuses on the gaze area. We process the head location map $\mathbf{H}$ through three max pooling operations followed by a flattening operation and then concatenate the output with the gaze-regression-specific feature $\mathbf{f}_{\rm gaze}^{\rm reg}$ to generate the attention map $\mathbf{A}^{'}$. It should be noted that since GaTector+ cannot obtain an accurate head location map during the training, we use a cross-entropy loss to supervise the learning of the attention map.

\noindent\textbf{Gaze prediction.}
After obtaining the attention map $\mathbf{A}$, we multiply it with the output of the scene feature module to obtain a scene feature that is aware of the gaze object:
\begin{equation}
    \mathbf{f}_{\rm scene}^{\rm attn} = \mathbf{f}_{\rm scene} \odot \mathbf{A}^{'},
\end{equation}
where $\odot$ represents the element-wise multiplication. In this way, the coarse-grained features containing the scene use attention to establish precise associations between the object and the target person. Then, we concatenate the gaze-aware scene feature with the gaze-regression-specific feature $\mathbf{f}_{\rm gaze}^{\rm reg}$ to generate the fused feature $\mathbf{f}_{\rm fused}$. Finally, we feed the fused feature $\mathbf{f}_{\rm fused}$ into the gaze predictor to predict a gaze heatmap $\mathbf{M}$. The gaze predictor is composed of an encoder-decoder structure and a $1\times1$ convolution layer. Specifically, We first compress the channels to 512 through the encoder and then use the decoder which consists of three deconvolution layers to restore the spatial resolution to the output size. Finally, we apply a $1\times1$ convolution layer on the output of the encoder-decoder to obtain a heatmap with a size of $64\times64$ as the prediction result. We follow Chong \etal~\cite{chong2020detecting} to calculate the gaze estimation loss $\mathcal{L}_{\rm gaze}$. Given the annotated gaze point $\mathbf{q} = (q_{x}, q_{y})$, we apply the Gaussian blur to generate the vanilla ground truth heatmap $\mathbf{T}\in \mathbb{R}^{H_{\rm T} \times W_{\rm T}}$. $H_{\rm T}$ and $W_{\rm T}$ represent the height and width of the heatmap.
\begin{equation}
    \begin{split}
        \widetilde{\mathbf{T}} = \frac{1}{2 \pi \sigma_{x} \sigma_{y} } \exp \left[-\frac{1}{2}\left(\frac{\left(x-q_{x}\right)^{2}}{\sigma_{x}^{2}}+\frac{\left (y-q_{y}\right)^{2}}{\sigma_{y}^{2}}\right)\right].
    \end{split}
    \label{eqGaussian}
\end{equation}
In ~\eqref{eqGaussian}, $\sigma_{x}$ and $\sigma_{y}$ indicate the standard deviation. We follow Chong \etal~\cite{chong2020detecting} and set $\sigma_{x} = 3$, $\sigma_{y} = 3$. Afterward, we normalize the heatmap and obtain the ground truth heatmap $\mathbf{T}_{i,j} = \frac{\widetilde{\mathbf{T}}_{i,j}}{{\rm max}(\widetilde{\mathbf{T}})}$. Given a predicted heatmap $\mathbf{M}\in \mathbb{R}^{H_{\rm T} \times W_{\rm T}}$, we calculate the mean square error to obtain the gaze estimation loss $\mathcal{L}_{gaze}$:
\begin{equation}
    \mathcal{L}_{\rm gaze} = \frac{1}{N_{\rm T}\times H_{\rm T} \times W_{\rm T}} \sum_{k=1}^{N_{\rm T}} \sum_{i=1}^{H_{\rm T}} \sum_{j=1}^{W_{\rm T}} (\textbf{M}_{k,i,j} - \textbf{T}_{k,i,j})^{2}.
    \label{eqLgaze}
\end{equation}
where N and k represent the number of people and the k-th person respectively. It should be noted that during the training process of our model, the $\mathcal {L}_{\rm gaze}$ is only calculated when the head location is detected. Specifically, GaTector+ returns the heatmap only when the head is present. Finally, the gaze heatmap is further guided by an energy aggregation loss to converge into the gaze object box. The energy aggregation loss is defined as:
\begin{equation}
    \mathcal{L}_{\rm gb} = {1} - \frac{1}{h \times w} \sum_{k=1}^{N} \sum_{i={x}_{1}}^{{x}_{2}} \sum_{j={y}_{1}}^{{y}_{2}} \textbf{M}_{i,j}.
    \label{eqenergy}
\end{equation}
where ${h},{w}$ presents the height and width of the ground truth gaze object box. By calculating the position of the maximum value of the gaze heatmap, we can easily obtain the gaze location.
 
\noindent\textbf{Attention supervision.} Different from the GaTector, GaTector+ does not have any head-related prior information, so it is difficult to obtain an accurate head location map during training, which makes it difficult to learn accurate attention weights. Therefore, we add supervision to the attention weights to get a more accurate attention distribution while accelerating the convergence of the model. Similar to the ground truth of gaze heatmap, we use a Gaussian blur matrix $\mathbf{A} \in \mathbb{R}^{N\times 7\times 7}$ centered on the gaze point as the ground truth of attention weights supervision. The difference is that we set the variance $\sigma$ of the Gaussian distribution to 0.35 to adopt the $7\times7$ attention weights matrix. Then we calculate the cross-entropy loss $\mathcal{L}_{attn}$ between $\mathbf{A}$ and the predicted attention weights matrix $\mathbf{A'}\in \mathbb{R}^{N\times 7\times 7}$:
\begin{equation}
    \begin{split}
        \mathcal{L}_{\rm attn} = -\frac{1}{N\times H\times W} \sum_{k=1}^{N} \sum_{i=1}^{H}  \sum_{j=1}^{W} a_{k,i,j} \log \hat{a}_{k,i,j}
    \end{split}
    \label{eqCrossenp}
\end{equation}
$a_{k,i,j}$ represents the value of the ground truth of each point in the attention weight matrix corresponding to the ${k}$-th person, and $\hat{a}_{k,i,j}$ represents the prediction weight. N, H, and W represent the number of people and the height and width of the attention weight matrix, respectively. The $\mathcal{L}_{attn}$ same as $\mathcal{L}_{gaze}$, it is only calculated when the head is present.


\begin{figure*}[!t]
\centering
\includegraphics[width=0.85\linewidth]{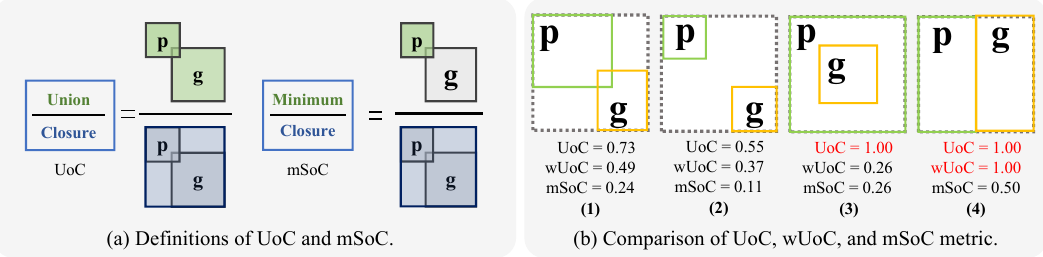}
\caption{(a) The definition of UoC and mSoC is demonstrated. (b) Comparison of the results of UoC, wUoC and mSoC in different situations: (1) and (2) are general situations; (3) For the ground truth box inside the prediction box, the UoC metric is invalid; (4) If the ground truth box and the prediction box have equal areas and are adjacent to each other, both UoC and wUoC are invalid.}
\vspace{-0.45cm}
\label{fig:msoc}
\end{figure*}


\subsection{Gaze object detection evaluation metric}
\label{sec:method_metric}
In this paper, we propose the mSoC metric for gaze object detection task, which is more reasonable and robust compared to the wUoC metric.

\noindent\textbf{Definition of wUoC.}
In Gatector~\cite{wang2022gatector}, the wUoC metric measures the distance between two boxes by the product of the UoC and a weight $\min{(\frac{p}{g}, \frac{g}{p})}$. 

\noindent\textbf{Limitations of wUoC.} 
We found some limitations to the wUoC metric. 
As shown in Fig~\ref{fig:msoc} (b) case (4), where two adjacent boxes have the same area, the value yielded by wUoC does not reflect the actual situation. Furthermore, although wUoC provides a higher value for case (2) than for case (3), it is evident that the error between the prediction and ground truth boxes is larger in case (2) than in case (3), its validity is compromised when the prediction box area is larger than the ground truth box and encompasses it. These cases illustrate that the wUoC metric does not accurately reveal the shape error of boxes.

\noindent\textbf{Definition of mSoC.}
We propose a new metric mSoC that can address these limitations produced by wUoC, as shown in Fig~\ref{fig:msoc} (a). Based on the UoC metric, our proposed mSoC is obtained by multiplying the UoC and a weight that calculates the difference between boxes and the minimum closure box.
\begin{equation}
    \text{mSoC} = \min{(\frac{p}{a}, \frac{g}{a})} \times \frac{{p} \cup {g}}{a}.
    \label{mSoC}
\end{equation}
where ${p}$ and ${g}$ are the predicted and ground truth boxes, respectively, and ${a}$ is their minimum closure bounding box.

\noindent\textbf{Advantages of mSoC.}
The mSoC metric is capable of revealing the difference between the prediction box and the ground truth box, not only the distance error but also the shape error in more complex scenarios. As shown in Fig~\ref{fig:msoc} (b), the wUoC metric is not sensitive enough to shape error and even fails in special cases such as (4). Our proposed mSoC can more accurately reveal the error of the boxes, which is because mSoC calculates the difference between boxes and the minimum closure box, it is more reasonable and robust. In \tabref{tab:metrics}, we verify our proposed mSoC metric by compared with wUoC in different thresholds.

\section{Experiments}
\label{sec:experiments}

\subsection{Experimental setting}
\label{sec:exp_settings}

\noindent\textbf{Dataset:} 
The GOO-synth and GOO-Real datasets are commonly used for gaze object detection tasks and are particularly challenging for small object detection, which contains 192,000 synthetic images and 9,552 real images, respectively. GOO-Synth and GOO-Real datasets contain annotations of gaze points, gaze objects, and bounding boxes for all objects and have 24 different categories. The GazeFollowing~\cite{recasens2015they} and VideoAttentionTarget~\cite{chong2020detecting} datasets are widely used in the training and verification of the gaze following task. GazeFollowing is a comprehensive image dataset comprising over 122,000 images in total, featuring annotations for more than 130,000 individuals. The test images are annotated with both gaze and head location information. VideoAttentionTarget dataset is a video dataset consisting of a wide range of video clips from YouTube, which consists of 1331 head tracks with 164,000 frame-level bounding boxes, and 109,574 gaze points.

\noindent\textbf{Metrics:} For the object detection performance, we use the average precision (AP) as our metric. We use AUC, L2 distance error (Dist.$\downarrow$), and angle error(Ang.$\downarrow$) to evaluate the performance of gaze prediction. AUC measures the area under the ROC curve, with higher values indicating better model performance. L2 distance bias (Dist.$\downarrow$) is used to measure the mean square distance between the predicted and ground truth heatmaps. The smaller the L2 distance, the more accurate the gaze point prediction. Angle error (Ang.$\downarrow$) is used to measure the angle between the center point of the predicted heatmap and the ground truth gaze point. For gaze object detection performance, we use the mSoC metric, which can reveal the differences between boxes and be more robust. For a fair comparison, when we evaluate the gaze following task on the GazeFollowing and VideoAttentionTarget datasets, we follow previous work~\cite{Tu_2022_CVPR, tu2023gaze} and use additional mAP to evaluate the comprehensive detection capabilities of the model, which includes head detection and gaze location prediction. In the process of calculating mAP, a predicted gaze instance detection is considered a true positive if and only if it is highly accurate (\ie, the IoU between the predicted human head box and ground truth is greater than 0.5 while the L2 distance is less than 0.15 and confidence for gaze object is larger than 0.75), and only calculate AUC and L2 distance for those predicted instances considered as true positives.

\noindent\textbf{Implementation details:} The backbone of our model uses ResNet-50~\cite{he2016deep} pre-trained on ImageNet. The network was optimized by Adam~\cite{kingma2014adam} for 100 epochs and initialized through normal initialization. For the first 50 epochs of training, we freeze the parameters of the backbone feature extraction network. We set the freeze batch size and unfreeze batch size as 32 and 16, respectively. The freeze initial learning rate was set as 1e-4, the unfreeze initial learning rate was set as 1e-4, and the learning rate is reduced by a factor of 0.94 every epoch. In ~\eqref{eqtotalloss}, we set $\alpha$ as 1.5. Following wang~\etal\cite{wang2022gatector}, we set $\beta$ and $\gamma$ as 10000 and 10 respectively. We set the input size as $224\times224$ and the predicted heatmap size as $64\times 64$. For a fair comparison, we are consistent with GaTector~\cite{wang2022gatector} and use YOLOv4~\cite{bochkovskiy2020yolov4} as the object detector for GaTector+. All experiments are implemented based on the PyTorch and one GeForce RTX 3090Ti GPU.

\subsection{Comparison with SOTA}
We compared the performance difference between GaTector+, GaTector-based head-free method, and head-required GaTector.
\begin{itemize}
\item \textbf{GaTector:} A unified framework for gaze object detection, which is a two-stage method. 
\item \textbf{GaTector w/o HB:} We remove the \textbf{head branch} in GaTector to implement the head-free method. 
\item \textbf{GaTector w/o HIM:} Since we only remove the head image input, the task-specific module may also help to improve the performance, we only remove the \textbf{head-input module} in GaTector.
\end{itemize}

\begin{table*}[!t]
  \centering
  \small
  \caption{Gaze object detection performance on the GOO-Synth dataset.}
  \setlength{\tabcolsep}{7pt}{
    \begin{tabular}{c|l|ccc|ccc|ccc}
    \toprule
    \multicolumn{2}{c|}{\multirow{2}[1]{*}{Method}} & \multicolumn{3}{c|}{gaze object detection} & \multicolumn{3}{c|}{Oject Detection} & \multicolumn{3}{c}{Gaze Estimation} \\
    \multicolumn{2}{c|}{} & mSoC  & mSoC$_{50}$ & mSoC$_{75}$ & AP    & AP$_{50}$  & AP$_{75}$  & AUC$\uparrow$  & Dist.$\downarrow$ & Ang.$\downarrow$ \\
    \midrule
    Head-required & GaTector~\cite{wang2022gatector} & 67.94 & 98.14 & 86.25 & 56.80  & 95.30  & 62.50  & 0.957  & 0.073  & 14.9  \\
    \midrule
    \multirow{3}[2]{*}{Head-free} & GaTector w/o HB & 55.01 & 92.70 & 60.81 & 41.51  & 90.73  & 31.34  & 0.933  & 0.130  & 24.9  \\
          & GaTector w/o HIM & 56.19 & 88.08 & 66.18 & 43.50  & 86.23  & 37.52  & 0.950  & 0.110  & 21.6  \\
          & GaTector+ & \textbf{67.25} & \textbf{97.94} & \textbf{84.12 } & \textbf{50.81 } & \textbf{96.29 } & \textbf{47.21 } & \textbf{0.951 } & \textbf{0.090 } & \textbf{17.9 } \\
    \bottomrule
    \end{tabular}%
    }
  \label{tab:sota_synth}%
\end{table*}%

\begin{table*}[!t]
  \centering
  \small
  \caption{Gaze object detection performance on the GOO-Real dataset.}
  \setlength{\tabcolsep}{7pt}{
    \begin{tabular}{c|l|ccc|ccc|ccc}
    \toprule
    \multicolumn{2}{c|}{\multirow{2}[2]{*}{Method}} & \multicolumn{3}{c|}{gaze object detection} & \multicolumn{3}{c|}{Object Detection} & \multicolumn{3}{c}{Gaze Estimation} \\
    \multicolumn{2}{c|}{} & mSoC  & mSoC$_{50}$ & mSoC$_{75}$ & AP    & AP$_{50}$  & AP$_{75}$  & AUC$\uparrow$  & Dist.$\downarrow$ & Ang.$\downarrow$ \\
    \midrule

\multicolumn{11}{l}{\textbf{Pre-trained on GOO-Synth}} \\
    \midrule
    Head-required & GaTector~\cite{wang2022gatector} & 71.24 & 97.51 & 88.78 & 58.23 & 96.82 & 64.40  & 0.940 & 0.087 & 14.8 \\
    \midrule
    \multirow{3}[2]{*}{Head-free} & GaTector w/o HB & 65.24 & 96.23 & 78.90 & 51.77  & 94.08  & 50.72  & 0.942 & 0.112 & 18.5 \\
          & GaTector w/o HIM & 66.40 & 96.22 & 80.49 & 52.76  & 94.65  & 53.50  & 0.941 & 0.103 & 17.0 \\
          & GaTector+ & \textbf{69.90} & \textbf{97.23} & \textbf{86.78} & ~\textbf{55.95 } & ~\textbf{96.28 } & ~\textbf{59.78 } & \textbf{0.953} & \textbf{0.084} & \textbf{14.5} \\
        \midrule
    \multicolumn{11}{l}{\textbf{No Pre-train}} \\
    \midrule
    Head-required & GaTector~\cite{wang2022gatector} & 62.42 & 95.11 & 73.54 & 49.85 & 93.11 & 48.86  & 0.927  & 0.196  & 39.5  \\
    \midrule
    \multirow{3}[1]{*}{Head-free} & GaTector w/o HB & 35.02 & 73.20 & 27.11 & 24.56 & 64.83 & 12.15 & 0.617  & 0.318  & 64.5  \\
          & GaTector w/o HIM & 40.87 & 82.05 & 33.54 & 29.21 & 74.02 & 16.13 & 0.661  & 0.332  & 59.8  \\
          & GaTector+ & \textbf{58.13} & \textbf{93.45} & \textbf{65.04} & \textbf{45.01} & \textbf{89.47} & \textbf{40.38} & ~\textbf{0.910 } & ~\textbf{0.126 } & ~\textbf{22.0 } \\

    \bottomrule
    \end{tabular}%
    }
  \label{tab:sota_real}%
\end{table*}%


\noindent\textbf{Experiments on GOO-Synth dataset.} We compare the GaTector and GaTector-based head-free methods with our proposed GaTector+. The performance of gaze object detection and gaze estimation on the GOO-Synth dataset can be observed in ~\tabref{tab:sota_synth} and ~\tabref{tab:ge_goosynth_real} respectively. 

The gaze object detection performance is shown in ~\tabref{tab:sota_synth}, the head-required method GaTector achieves the best result of 67.92\% mSoC. To explore the head-free method, we first attempted to directly remove the head branch of GaTector, which achieves the result of 55.01\% mSoC and is significantly lower than the performance of GaTector. This suggests that head prior information is important for gaze object detection and that the task-specific module may also help to improve performance. Under this inspiration, we attempt to extract gaze-specific features from scene images by only removing the head-input module of GaTector, which achieves better results than removing the head branch. Our proposed GaTector+ results are much higher than other head-free settings. Compared to the head-required method, our GaTector+ is only 0.69\% (67.25\%~\vs~67.94\%) lower than GaTector in terms of mSoC. 

\begin{table*}[!t]
  \centering
  \small
  \caption{Gaze following results on the GOO-Synth dataset and GOO-Real datasets.}
  \setlength{\tabcolsep}{7pt}{
    \begin{tabular}{c|l|ccc|ccc|ccc}
    \toprule
    \multicolumn{2}{c|}{\multirow{2}[4]{*}{Method}} & \multicolumn{3}{c|}{GOO-Synth} & \multicolumn{3}{c|}{GOO-Real Pre-train} & \multicolumn{3}{c}{GOO-Real No Pre-train} \\
\cmidrule{3-11}    \multicolumn{2}{c|}{} & AUC$\uparrow$  & Dist.$\downarrow$ & Ang.$\downarrow$  & AUC$\uparrow$  & Dist.$\downarrow$ & Ang.$\downarrow$  & AUC$\uparrow$  & Dist.$\downarrow$ & Ang.$\downarrow$ \\
    \midrule
    \multicolumn{1}{c|}{\multirow{5}[2]{*}{Head-required}} & Random & 0.497  & 0.545  & 77.0  & 0.512  & 0.453  & 78.9  & 0.501  & 0.511  & 79.1  \\
          & Recasens~\cite{recasens2015they} & 0.929  & 0.162  & 33.0  & 0.903  & 0.195  & 39.8  & 0.850  & 0.220  & 44.4  \\
          & Lian~\cite{lian2018believe}  & 0.954  & 0.107  & 19.7  & 0.890  & 0.168  & 32.6  & 0.840  & 0.321  & 43.5  \\
          & Chong\cite{chong2020detecting} & 0.952  & 0.075  & 15.1  & 0.889  & 0.150  & 29.1  & 0.796  & 0.252  & 51.4  \\
          & GaTector~\cite{wang2022gatector} & \textbf{0.957}  & \textbf{0.073}  & \textbf{14.9}  & \textbf{0.940}  & \textbf{0.087}  & \textbf{14.8}  & \textbf{0.927}  & \textbf{0.196}  & \textbf{39.5}  \\
    \midrule
    \multicolumn{1}{c|}{\multirow{3}[2]{*}{Head-free}} & GaTector w/o HB & 0.933  & 0.130  & 24.9  & 0.942  & 0.112  & 18.5  & 0.617  & 0.318  & 64.5  \\
          & GaTector w/o HIB & 0.950  & 0.110  & 21.6  & 0.940  & 0.103  & 17.0  & 0.661  & 0.332  & 59.8  \\
          & GaTector+ & \textbf{0.951}  & \textbf{0.090}  & \textbf{17.9}  & ~\textbf{0.953 } & ~\textbf{0.084 } & ~\textbf{14.5 } & \textbf{0.910}  & ~\textbf{0.126 } & ~\textbf{22.0 } \\
    \bottomrule
    \end{tabular}%
    }
  \label{tab:ge_goosynth_real}%
\end{table*}%

The gaze estimation performance is shown in ~\tabref{tab:ge_goosynth_real}, compared to previous head-required methods~\cite{recasens2015they, lian2018believe, chong2020detecting} on the GOO-Synth dataset, GaTector achieved the best performance for AUC (0.957), L2 distance error (0.073), and angle error ($14.9^{\circ}$).  In the head-free method, GaTector w/o HB achieves the result for AUC (0.933), L2 distance error(0.130), and angle error ($24.9^{\circ}$). Since the head branch is crucial for gaze estimation, we remove the head-input module of GaTector. This modification led to a significant improvement in performance. Our GaTector+ achieves the result for AUC (0.951), L2 distance error(0.090), and angle error ($17.9^{\circ}$), which is much higher than GaTector without the head branch. It is worth noting that the performance of our GaTector+ has surpassed some head-required methods~\cite{recasens2015they,lian2018believe} in the 
L2 distance and angle error.

The experimental results demonstrate that our GaTector+ achieves state-of-the-art on head-free gaze object detection and gaze following tasks. The main reasons for the performance improvement are as follows: \textbf{(1)} Our SGS+ can extract more accurate gaze-related features based on scene images, and at the same time use attention to establish accurate associations between objects and target person, thus obtaining more accurate prediction results. \textbf{(2)} Our supervision of the attention loss enables the model to obtain more accurate attention, which in turn guides the gaze predictor to regress a more accurate gaze heatmap. \textbf{(3)} The head location map provided by the head detector and joint optimization by the box aggregation energy loss contribute significantly to the performance improvement.

\begin{table}[!t]
  \centering
  \small
  \caption{Gaze following results on GazeFollowing and VideoAttentionTarget datasets. For a fair comparison, we use the same evaluation method as GTR~\cite{tu2023gaze}.}
  \setlength{\tabcolsep}{2pt}
  
    \begin{tabular}{l|cccc|ccc}
    \toprule
    \multicolumn{1}{c|}{\multirow{2}[4]{*}{Method}} & \multicolumn{4}{c|}{GazeFollowing} & \multicolumn{3}{c}{VideoAttentionTarget} \\
\cmidrule{2-8}          & AUC$\uparrow$   & \tabincell{c}{Min \\ Dist.}$\downarrow$ & \tabincell{c}{Avg \\ Dist.}$\downarrow$ & mAP$\uparrow$   & AUC$\uparrow$   & \tabincell{c}{L2 \\ Dist.}$\downarrow$ & mAP$\uparrow$ \\
    \midrule
    Random & 0.391  & 0.533  & 0.487  & 0.104  & 0.247 & 0.592 & 0.091 \\
    Center & 0.446  & 0.495  & 0.371  & 0.117  & -     & -     & - \\
    Fixed bias & -     & -     & -     & -     & 0.522 & 0.472 & 0.130  \\
    Recasens~\cite{recasens2015they} & 0.804  & 0.233  & 0.124  & 0.457  & -     & -     & - \\
    Chong~\cite{chong2018connecting} & 0.807  & 0.207  & 0.120  & 0.449  & 0.791 & 0.214 & 0.374 \\
    Lian~\cite{lian2018believe}  & 0.881  & 0.153  & 0.087  & 0.469  & 0.784 & 0.172 & 0.392 \\
    Chong~\cite{chong2020detecting} & 0.902  & 0.142  & 0.082  & 0.483  & 0.812 & 0.146 & 0.420 \\
    HGTTR~\cite{Tu_2022_CVPR} & 0.917  & 0.069  & 0.133  & 0.547  & 0.893 & 0.137 & 0.514 \\
    GTR~\cite{tu2023gaze}   & 0.928  & 0.057  & 0.114  & 0.604  & 0.925 & 0.093 & 0.584 \\
    \midrule
    GaTector~\cite{wang2022gatector} & 0.896  & 0.122  & 0.209  & -     & 0.881 & 0.175 & - \\
    GaTector+ & \textbf{0.931 } & \textbf{0.047 } & 0.131  & \textbf{0.679 } & \textbf{0.944} & \textbf{0.068} & \textbf{0.598} \\
    \bottomrule
    \end{tabular}%
  \label{tab:gaze_location}%
\end{table}%

\noindent\textbf{Experiments on GOO-Real dataset.} We also report the gaze object detection and gaze estimation performance by comparing the GaTector and GaTector-based head-free methods with our proposed GaTector+ on the GOO-Real dataset, which is summarized in ~\tabref{tab:sota_real} and ~\tabref{tab:ge_goosynth_real}. Since the GOO-Real dataset contains a limited number of images, we initially report the results obtained by training a model that was pre-trained on the GOO-Synth dataset. To demonstrate our models' robustness, we also compare each model's performance without pre-training.

~\tabref{tab:sota_real} shows the performance of gaze object detection with and without pre-training. Compared to the GaTector-based head-free methods, our GaTector+ achieves the best results (69.90\% mSoC) when the model is pre-trained on the GOO-Synth dataset, which is higher than all GaTector-based head-free methods and already very close to GaTector's 71.24\%. When our GaTector+ was not pre-trained on the GOO-Synth dataset, its performance on the GOO-Real dataset was relatively worse compared to when it was pre-trained, but it still outperforms all GaTector-based head-free methods and is close to GaTector (58.13\% \vs 62.42\%)

\tabref{tab:ge_goosynth_real} report the gaze estimation performance. Considering that the GOO-Real dataset contains a limited number of images, which may cause some models to not converge, we first pre-trained on the GOO-Synth dataset in advance and then trained on the GOO-Real dataset for a fair comparison. Among all head-required methods, our GaTector achieves the best performance on all three metrics. Our GaTector+ achieves the best performance compared with head-free methods and even surpasses all head-required methods~\cite{recasens2015they,lian2018believe,chong2020detecting,wang2022gatector} in all three metrics. We also report the gaze estimation performance of the model on the GOO-Real dataset without pre-training. We can observe that GaTector+ shows an obvious advantage when there is no pre-training.

The above experiments indicate that our GaTector+ has a strong generalization ability and can continue to optimize real images based on features extracted from synthetic images. 

\noindent\textbf{Experiments on GazeFollowing and VideoAttentionTarget datasets.} In~\tabref{tab:gaze_location}, we further verified the performance of our model on the gaze following task on the GazeFollowing and VideoAttentionTarget datasets. Compared to GTR, our model’s mAP increased by 7.5\%, which shows that our model can detect more effective prediction instances under the same conditions, including human heads and gaze heatmaps. On the VideoAttentionTarget dataset, although our model and GTR have only a slight advantage in the mAP metric, our L2 distance is reduced by 0.025 and AUC has also increased significantly. These results show that compared to HGTTR~\cite{Tu_2022_CVPR} and GTR~\cite{tu2023gaze} using Transformer for global modeling, our head-based attention and attention supervision mechanism can search the visual attention focus based on the head more effectively, thereby regressing a more accurate gaze heatmap for visual focus.


\begin{table*}[!t]
  \centering
  \small
  \caption{Ablation comparison about each component on the GOO-Synth dataset.}
  \setlength{\tabcolsep}{5pt}
    \begin{tabular}{cl|ccc|ccc|ccc}
    \toprule
    \multicolumn{2}{c|}{\multirow{2}[1]{*}{Setups}} & \multicolumn{3}{c|}{Gaze Object Detection} & \multicolumn{3}{c|}{Object Detection} & \multicolumn{3}{c}{Gaze Estimation} \\
    \multicolumn{2}{c|}{} & mSoC  & \multicolumn{1}{c}{mSoC$_{50}$} & \multicolumn{1}{c|}{mSoC$_{75}$} & AP    & AP$_{50}$  & AP$_{75}$  & AUC$\uparrow$  & Dist.$\downarrow$ & Ang.$\downarrow$ \\
    \midrule
        \romannumeral1  & GaTector+ w/o input-specific &  65.35     &   97.33    &    80.16   &    50.03   &    95.85   &    44.91   &    0.949   &   0.120    & 23.7 \\
        \romannumeral2  & GaTector+ w/o gaze-related specific &    65.53   &   97.18    &    81.38   &    50.59   &   95.75    &    47.07   &    0.943   &    0.094   & 19.1 \\
        \romannumeral3  & GaTector+ w/o gaze-specific &  65.22     &    97.50   &    80.45   &    49.27   &   95.78    &    43.92   &    0.944   &    0.096   & 19.6 \\
        \romannumeral4  & GaTector+ w/o scene-specific &   64.70    &    96.92   &    79.91   &    49.34   &   95.32    &    44.06   &    \textbf{0.953}   &   0.099    &  19.8 \\
        \romannumeral5  & GaTector+ w/o defocus layer &   56.86    &   88.96    &    66.52   &   41.82    &  83.94  &    36.56   &    0.946   &  0.095     &  18.2 \\
        \romannumeral6  & GaTector+ w/o head defocus &   65.81    &    97.45   &   82.01    &    50.15   &    95.89   &    45.88   &    0.951   &    0.096   &  20.6 \\
        \romannumeral7  & GaTector+ w/o $\mathcal{L}_{eng}$ & 64.74 & 97.01 & 79.97 & 49.36  & 95.29  & 44.08  & 0.940  & 0.092  & 18.4  \\
        \romannumeral8  & GaTector+ w/o $\mathcal{L}_{attn}$ & 65.43 & 97.23 & 80.87 & 50.56  & 95.54  & \textbf{47.59}  & 0.947  & 0.097  & 19.9  \\
    \midrule
        \romannumeral9  & GaTector+ & \textbf{67.25} & \textbf{97.94} & ~\textbf{84.12 } & ~\textbf{50.81 } & ~\textbf{96.29 } & 47.21  & 0.951 & ~\textbf{0.090 } & ~\textbf{17.9 } \\
    \bottomrule
    \end{tabular}%
  \label{tab:ablation}%
\end{table*}%

\subsection{Ablation Studies}
\label{sec:exp_ablation}

\noindent\textbf{Ablation study about each component.} 
We reported the ablation study results on the GOO-Synth dataset in ~\tabref{tab:ablation} to analyze the impact of each component of our proposed GaTector+.

\begin{itemize}
\item \textbf{\romannumeral1~GaTector+ w/o input-specific.} When we remove the input-specific blocks and directly input the image into the backbone, we observed that the performance dropped sharply, especially gaze estimation, which demonstrates the necessity of using input-specific blocks to transform the original image into the general feature space before feeding the image into the backbone.
\item \textbf{\romannumeral2~GaTector+ w/o gaze-related specific.} When the two gaze-related specific convolution blocks after the backbone are removed at the same time, the performance of gaze estimation drops significantly, which indicates that gaze-related specific features are very important for gaze estimation.

\item \textbf{\romannumeral3~GaTector+ w/o gaze-specific.} After removing the gaze-specific convolutional blocks, the performance of gaze estimation drops to varying degrees on all three metrics, which proves the necessity of our extraction of gaze-specific features.

\item \textbf{\romannumeral4~GaTector+ w/o scene-specific.} Similar to the result obtained after removing the gaze-specific convolutional block, the performance of gaze estimation also drops after removing the scene-specific convolutional block, indicating that salient features from the scene can also bring gains to gaze estimation.

\item \textbf{\romannumeral5~GaTector+ w/o defocus layer.} We observe a significant drop in object detection performance when we remove the defocus layer in the head-specific block and object-specific block, which further results in a decline in gaze object detection performance. Such experimental results show that the defocus layer is very effective and efficient for object detection in small and dense retail scenes.

\item \textbf{\romannumeral6~GaTector+ w/o detection-head defocus.} We observe a significant drop in object detection performance when using upsample to replace the defocus detection head in the objector detector. This indicates that the defocus detection head in the object detector has a positive effect on improving the performance of small object detection in retail scenarios.

\item \textbf{\romannumeral7~GaTector+ w/o $\mathcal{L}_{eng}$.} Without box aggregation energy loss supervision, the performance of gaze object detection decreased by 2.51\% mSoC, and the performance of gaze estimation and object detection also decreased to varying degrees due to the lack of joint optimization. This shows that the box aggregation energy loss plays a role in the joint optimization of object detection and gaze estimation.

\begin{table}[!t]
\small
\vspace{-0.2cm}
  \centering
  \caption{Comparison of wUoC and mSoC on the GOO-Synth dataset.}
  \setlength{\tabcolsep}{1.10pt}
    \begin{tabular}{l|rrrrrrrrrr}
    \toprule
    metric & \multicolumn{1}{c}{0.5} & \multicolumn{1}{c}{0.55} & \multicolumn{1}{c}{0.6} & \multicolumn{1}{c}{0.65} & \multicolumn{1}{c}{0.7} & \multicolumn{1}{c}{0.75} & \multicolumn{1}{c}{0.8} & \multicolumn{1}{c}{0.85} & \multicolumn{1}{c}{0.9} & \multicolumn{1}{c}{0.95} \\
    \midrule
   wUoC  & 90.13  & 90.07  & 89.97  & 89.82  & 89.43  & 88.47  & 86.34  & 80.58  & 63.56  & 24.14  \\
    mSoC  & 97.94  & 97.67  & 97.05  & 95.73  & 92.40  & 84.12  & 65.91  & 35.08  & 6.55  & 0.08  \\
    \bottomrule
    \end{tabular}%
  \label{tab:metrics}%
\end{table}%

\item \textbf{\romannumeral8~GaTector+ w/o $\mathcal{L}_{attn}$.} After removing attention loss, the performance of gaze estimation drops significantly. This indicates that through the supervision of attention loss, the model can provide a more accurate attention map, thus guiding the gaze predictor to obtain a more accurate gaze heatmap.

\end{itemize}

\begin{table}[!t]
  \centering
  \small
  \caption{Gaze prediction results on the GOO-Synth dataset using different loss functions to supervise attention weights.}
  \setlength{\tabcolsep}{14.2pt}
    \begin{tabular}{l|ccc}
    \toprule
    \multicolumn{1}{c|}{Loss} & AUC$\uparrow$  & Dist.$\downarrow$ & Ang.$\downarrow$ \\
    \midrule
    MSE   & 0.944  &  0.094  &  19.7 \\
    Cross-Entropy & \textbf{0.951}   &   \textbf{0.090}    &  \textbf{17.9}  \\
    \bottomrule
    \end{tabular}%
  \label{tab:attention_loss}%
\end{table}%

\noindent \textbf{mSoC~\vs ~wUoC}
\tabref{tab:metrics} report wUoC and mSoC results of GaTector+ at different thresholds on the GOO-Synth dataset, respectively. We can observe that wUoC maintains a relatively high range when the threshold is greater than 0.7. The reason for this phenomenon is that the existence of such conditions in Fig~\ref{fig:msoc}~(b) causes the wUoC results to be abnormal.  Especially when the threshold is equal to 0.95, wUoC obtains a result of 24.14\%, which is unreasonable. Our proposed mSoC will drop sharply as the threshold is greater than 0.75. We believe that the main reason for this result is that: \textbf{(1)} In retail scenes with small and dense objects, wUoC tends to exhibit more situations like the one shown in~\tabref{fig:msoc}~(b)~(4), where two boxes are close to each other. This results in the false positive box being counted as true positive, further leading to wUoC maintaining a relatively high level even under a high threshold. \textbf{(2)} In contrast, our proposed mSoC can well handle the case where two boxes are adjacent, avoiding the error similar to wUoC, and thus obtaining a more reasonable result, making mSoC more restrictive.

\vspace{-0.2cm}
\subsection{Model Analysis}

\noindent\textbf{Analysis of model parameters and computational cost.} 

\begin{table*}[!t]
  \centering
  \small
  \caption{Analysis of model parameters and computational cost. We explored the parameters and computational cost of the model under different settings.}
  \setlength{\tabcolsep}{5pt}
    \begin{tabular}{cl|ccc|ccc|ccc|cc}
    \toprule
    \multicolumn{2}{c|}{\multirow{2}[1]{*}{Setups}} & \multicolumn{3}{c|}{Gaze Object Detection} & \multicolumn{3}{c|}{Object Detection} & \multicolumn{3}{c|}{Gaze Estimation} & \multirow{2}[1]{*}{Para.} & \multirow{2}[1]{*}{FLOPs} \\
    \multicolumn{2}{c|}{} & mSoC  & mSoC$_{50}$ & mSoC$_{75}$ & AP    & AP$_{50}$  & AP$_{75}$  & AUC$\uparrow$  & Dist.$\downarrow$ & Ang.$\downarrow$   &       &  \\
    \midrule
      \romannumeral1    & Low resolution & 56.86  & 88.96  & 66.52  & 41.82  & 83.94  & 36.53  & 0.946  & 0.095  & 18.2  & 75.53  & 25.81 \\
      \romannumeral2    & Interpolation & 65.14  & 97.27  & 80.95  & 49.92  & 94.74  & 46.29  & 0.954  & 0.096  & 19.2  & 77.39  & 26.31 \\
      \romannumeral3    & Crop gaze-specific feature & 65.56  & 97.71  & 82.07  & 49.56  & 96.19  & 44.07  & 0.940  & 0.095  & 20.3  & 71.13  & 21.88 \\
      \romannumeral4    & Gaze-scene seprate & 68.51  & 98.74  & 86.75  & 52.89  & 97.81  & 49.59  & 0.954  & 0.094  & 18.7  & 99.24  & 26.22 \\
      \midrule
      \romannumeral5    & GaTector+  & 67.25  & 97.94  & 84.12  & 50.81  & 96.29  & 47.21  & 0.951  & 0.090  & 17.9  & 75.73  & 26.22 \\
    \bottomrule
    \end{tabular}%

  \label{tab:model_analysis}%
\end{table*}%

\begin{itemize}
    \item \textbf{Analysis of the object-specific feature.} In \tabref{tab:model_analysis} \romannumeral1, if the original ResNet-50 features are used to perform object detection, their relatively small spatial resolution will lead to a sharp drop in detection performance. Although using the original features of ResNet-50 have a slight advantage in terms of computational cost and parameters, in retail scenes where objects are small and dense, larger feature maps are necessary for better object detection performance. In \tabref{tab:model_analysis} \romannumeral2, we use the interpolation to enlarge the feature map, and we observe that the object detection performance has a significant increase compared to the low-resolution feature map (41.82\% ~\vs~49.92\% ). Although enlarging the feature map will sacrifice parameters and computational costs, the gain it brings is very obvious. Our GaTector+ uses the defocus layer to enlarge the feature map. Compared with interpolation, it further improves the performance and reduces the model parameters and computational costs.

    \item \textbf{Analysis of the gaze-specific feature.} In \tabref{tab:model_analysis} \romannumeral3, in order to further reduce the parameters and computational cost of the model, we explored directly clipping gaze-specific features from the scene branch according to the detected head bounding box. Although the parameters and calculation costs have been significantly reduced, the gaze estimation performance also drops, which shows the necessity of the existence of the gaze branch, which can bring obvious performance increase to gaze estimation.
    
    \item \textbf{Analysis of the shared backbone.} In \tabref{tab:model_analysis} \romannumeral4, We use two backbones to extract gaze features and scene features respectively. Although this has achieved the best results, the parameters of the model have risen sharply, while our GaTector+ can still maintain a relatively low parameter and computational cost without significant performance degradation. 
\end{itemize}

\begin{figure}[!t]
\centering
\includegraphics[width=1\linewidth]{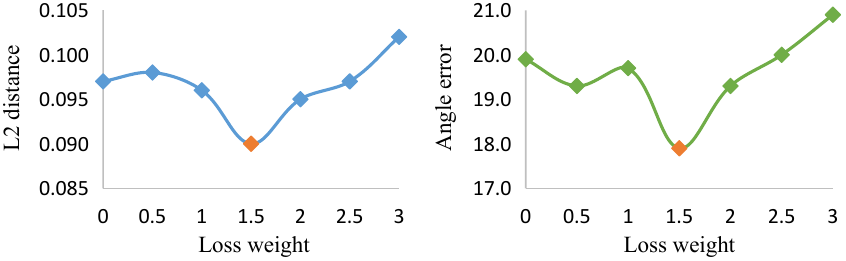}
\caption{Effect of attention loss hyperparameter on gaze estimation. The hyperparameter is varied from 0 to 3 with a step size of 0.5. The blue line represents the L2 distance, while the green line represents the angle error. The results indicate that the model achieves the best performance when the loss weight is equal to 1.5.}
\vspace{-0.45cm}
\label{fig:loss_hyper}
\end{figure}

\noindent\textbf{Analysis of the attention supervision.} 
~\tabref{tab:attention_loss} analyzes the gaze estimation performance of GaTector+ under different loss function supervision on the GOO-Synth dataset. Inspired by the gaze regressor, we first supervise the attention weights using Mean Squared Error (MSE), which resulted in an AUC score of 0.944, an L2 distance error of 0.094, and an angle error of $19.7^{\circ}$. When we use cross-entropy as the supervision method, it is more effective than MSE, resulting in an AUC score of 0.951, an L2 distance error of 0.090, and an angle error of $17.9^{\circ}$. This is mainly because the attention weights generated early in the model provide relatively ambiguous information. Using MSE loss to supervise attention weights makes it difficult for the model to converge. However, cross-entropy loss combined with the softmax activation function can accelerate the training speed of the model and better fit the distribution of attention weights. In order to weigh the proportion of each task and explore the optimal performance of the model, we adjusted the weight of the attention loss (as shown in ~\figref{fig:loss_hyper}). After adjustment, we finally determined that the model achieved the best performance when the attention loss hyperparameter was set to 1.5.

\begin{table}[!t]
  \centering
  \small
  \caption{Analysis of the bottleneck of GaTector and GaTector+ on the GOO-Synth.}
  \setlength{\tabcolsep}{8pt}
    \begin{tabular}{cc|cc|cc}
    \toprule
    \multicolumn{2}{c|}{Gaze Heatmap} & \multicolumn{2}{c|}{Object Box} & \multicolumn{2}{c}{Gaze Object Detection} \\
\cmidrule{1-6}    GT    & Pred. & GT    & Pred. & \multicolumn{1}{l}{GaTector} & \multicolumn{1}{l}{GaTector+} \\
\midrule
          & $\checkmark$ &       & $\checkmark$ & 67.94  & 67.25 \\
          & $\checkmark$ & $\checkmark$ &       & 70.99 & 69.74 \\
    $\checkmark$ &       &       & $\checkmark$ & \textbf{72.34} &  \textbf{72.58}\\
    \bottomrule
    \end{tabular}%
  \label{tab:bottleneck}%
\end{table}%

\noindent\textbf{Analysis of the bottleneck.} 
~\tabref{tab:bottleneck} analyzes the performance bottleneck of GaTector and our proposed GaTector+. Using predicted gaze heatmap and object bounding boxes, GaTector gets {67.94\%} mSoC. Our GaTector+ can achieve 67.25\% mSoC during the inference. 
After replacing the predicted results of the object detector with ground truth bounding boxes, the gaze object detection performance of GaTector+ can be improved to 69.74\% mSoC, which is only 2.49\% mSoC higher than the results predicted with the use of an object detector. GaTector can be improved to {70.99\%} in the same setting. This verifies the current approach has achieved accurate object detection performance.
When we replace the prediction results of the gaze regressor with the ground truth gaze heatmap, the GaTector+ performance of the gaze object detection will greatly improve by 5.33\% mSoC (72.58~\vs~67.25). GaTector performance can be improved to {72.34\%}. Therefore, we believe that the current bottleneck of gaze object detection is mainly on gaze regressor. This performance bottleneck analysis provides a promising direction for future research.

\begin{figure*}[!t]
\centering
\includegraphics[width=0.9\linewidth]{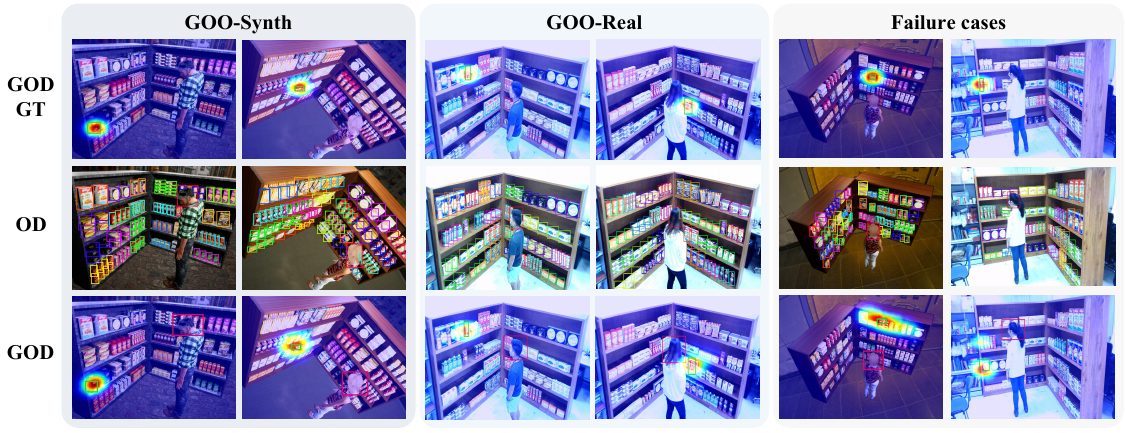}
\caption{Head-free gaze object detection visualization results of GaTector+ on the GOO-Synth and GOO-Real datasets. We provide two failure cases. The red bounding box represents the head of the target person, and the green bounding box represents the gazed object corresponding to the head.}
\label{fig:visualization}
\end{figure*}

\begin{figure*}[!t]
\centering
\includegraphics[width=0.9\linewidth]{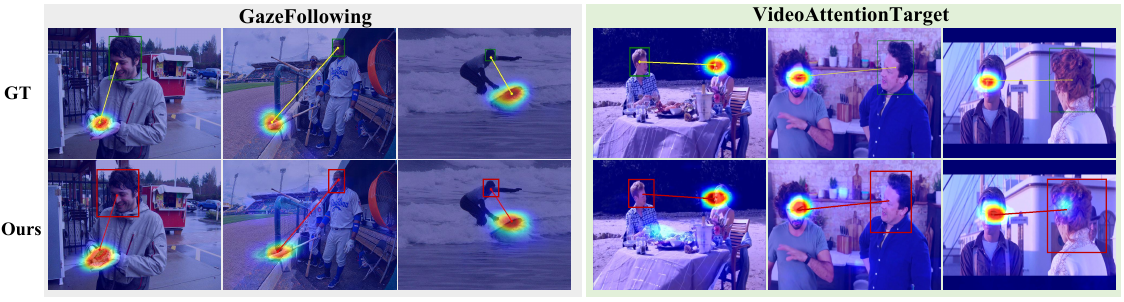}
\caption{Gaze following visualization results of GaTector+ on the GazeFollowing and VideoAttentionTarget datasets. We show the results of ground truth (GT) and ours gaze following.}
\label{fig:exted_visualization}
\end{figure*}

\begin{figure}[!t]
\centering
\includegraphics[width=0.9\linewidth]{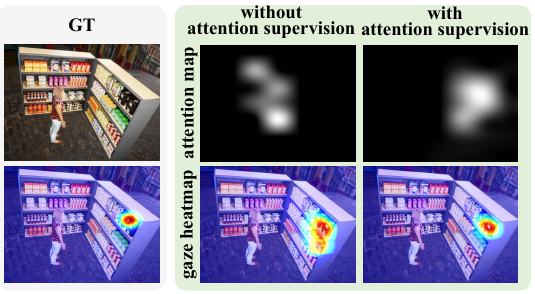}
\caption{Comparison of the attention map and predicted gaze heatmap on the GOO-Synth dataset under the supervision of attention loss. After adding attention loss, more accurate attention is obtained, thus obtaining a heatmap closer to the ground truth.}
\vspace{-0.45cm}
\label{fig:visualization_attention}
\end{figure}

\section{Visualization}
\label{sec:visualization}
\noindent\textbf{Visualization about gaze object detection.} Fig.~\ref{fig:visualization} presents qualitative results on the GOO-Synth dataset (the left four columns) and the GOO-Real dataset (the right three columns) for the proposed GaTector+. Our model is capable of accurately localizing item bounding boxes and relatively precisely predicting gaze heatmaps, resulting in accurate gaze object detection results in various retail scenarios (\eg different distances to retail objects, shooting views, and the number of people in the same image). However, there are some failure cases where errors in the gaze heatmap mislead the gaze object detection process, for the GOO-Synth and GOO-Real datasets, respectively. This demonstrates the direct impact of gaze estimation results on gaze object detection.

\noindent\textbf{Visualization about gaze following.} Fig.~\ref{fig:exted_visualization} shows the gaze following visualization results on GazeFollowing and VideoAttentionTarget datasets. Our method enables accurate head detection and regresses accurate gaze heatmaps in a variety of scenarios, which demonstrate the effectiveness of our model on gaze following tasks and can effectively acquire visual focus, resulting in a more comprehensive understanding of the gaze environment.


\noindent\textbf{Visualization about attention map.} In Fig.~\ref{fig:visualization_attention}, we provide visualization of the attention map and gaze heatmap predicted by GaTector+ without attention loss supervision, which shows relatively poor results. In particular, the resulting attention map does not provide useful information. When using attention loss supervision, GaTector+ can predict a high-quality attention map, and regress a more accurate gaze heatmap. This verifies that our proposed attention supervision has a positive effect on generating an accurate attention map.

\section{Conclusion}
\label{sec:conclusion}
This paper proposes GaTector+, a unified head-free framework for gaze object detection and gaze following, which does not need any auxiliary detector network to extract head-related prior information to predict the gaze object and visual focus of the target person in the image. To establish a connection between scene information and gaze information
in a unified framework without head-prior information, we
extend the SGS mechanism and extract head-specific features
from the scene image through a head-specific block. To obtain head-related information without prior knowledge, we embed a head detector to detect the head location, then a head-based attention mechanism is proposed to introduce gaze contexts to the scene-specific features. To mitigate the learning difficulty of the gaze regressor due to the inaccurate predicted head location, we propose an attention supervision mechanism to accelerate the learning process of the gaze regressor. Finally, to more accurately evaluate the performance of the gaze object detection task, we propose a novel evaluation metric mean Similarity over Candidates (mSoC), which can accurately evaluate gaze object detection in various situations. Experiments on multiple benchmark datasets demonstrate the superiority of our model on gaze object detection and gaze following tasks when head priors are not provided.


\bibliographystyle{IEEEtran}
\bibliography{egbib}


\end{document}